# Fish Mouth Inspired Origami Gripper for Robust Multi-Type Underwater Grasping

Honghao Guo, Junda Huang, Ian Zhang, Boyuan Liang, Xin Ma, *Member, IEEE,*
Yunhui Liu, *Fellow, IEEE,* and Jianshu Zhou, *Member, IEEE*

*Abstract*—Robotic grasping and manipulation in underwater environments present unique challenges for robotic hands traditionally used on land. These challenges stem from dynamic water conditions, a wide range of object properties from soft to stiff, irregular object shapes, and varying surface frictions. One common approach involves developing finger-based hands with embedded compliance using underactuation and soft actuators. This study introduces an effective alternative solution that does not rely on finger-based hand designs. We present a fish mouth inspired origami gripper that utilizes a single degree of freedom to perform a variety of robust grasping tasks underwater. The innovative structure transforms a simple uniaxial pulling motion into a grasping action based on the Yoshimura crease pattern folding. The origami gripper offers distinct advantages, including scalable and optimizable design, grasping compliance, and robustness, with four grasping types: pinch, power grasp, simultaneous grasping of multiple objects, and scooping from the seabed. In this work, we detail the design, modeling, fabrication, and validation of a specialized underwater gripper capable of handling various marine creatures, including jellyfish, crabs, and abalone. By leveraging an origami and bio-inspired approach, the presented gripper demonstrates promising potential for robotic grasping and manipulation in underwater environments.

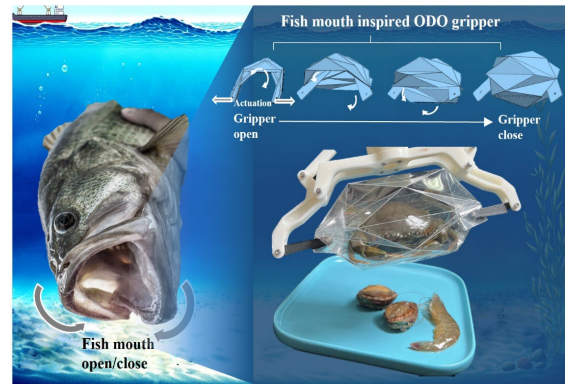

**Figure 1**. The proposed origami gripper inspired by the fish mouth successfully and robustly catches a crab.

## I. INTRODUCTION

Despite covering 70.8% of the Earth's surface, the vast expanse of the ocean remains largely unexplored [1]. To date, less than 5% of the world's oceans have been meaningfully studied, leaving the majority of marine species undiscovered. Aquaculture also plays a vital role in ensuring global food security by providing a sustainable source of nutritious seafood and supplementing dwindling wild-caught fish populations [2]. Additionally, salvaging ancient ruins and cultural relics submerged in the sea offers significant historical and cultural value, despite the challenges posed by marine factors such as strong currents and corrosive saltwater [3, 4]. These points underscore the importance of marine exploration and sampling [5].

The discrepancy in marine exploration is primarily due to limited access to marine environments and the challenges associated with sampling and studying marine organisms. The lack of comprehensive sampling tools capable of efficiently capturing and studying elusive marine organisms has hindered our understanding of their biology, behavior, and ecological roles. Traditional sampling and aquaculture tools face several disadvantages, including size and depth restrictions as well as a heavy reliance on skilled labor. Robotic sampling and manipulation solutions offer a promising avenue for enhancing underwater applications and explorations [6, 7].

While robotic grippers have been developed for terrestrial use, addressing these challenges underwater remains difficult. Rigid robot hands are suitable for handling solid targets like rocks but may harm fragile marine animals. Additionally, grasping planning becomes complicated in underwater dynamic situations, especially with irregularly shaped objects and varying surface frictions. This necessitates the development of specialized robotic grippers that can operate effectively in underwater environments while minimizing harm to delicate marine life [8].

Built upon classical finger-based designs, underactuated grippers utilizing underactuated mechanisms and soft grippers based on soft actuators have proven effective for underwater applications [9-12]. While soft robotic hands are viable, they struggle with sealing under high pressure and maintaining robust pressure control [13, 14]. Fluid-driven soft hands are safe and dexterous for land grasping, towards food even for delicate tofu [15, 16], rigid and sharp surface durian [17], and achieving dexterous in-hand manipulation [18, 19], however, the soft actuator faces challenge in underwater application in

* This work was supported in part by the Research Grants Council of Hong Kong under Grant 14204423 and 14207423.

Honghao Guo, Junda Huang, Xin Ma and Yunhui Liu are with the Department of Mechanical and Automation Engineering, The Chinese University of Hong Kong.

Ian Zhang, Boyuan Liang and Jianshu Zhou are with the Department of Mechanical Engineering, University of California, Berkeley.

terms of sealing, pressure resistance, and accompanied actuation [20-22]. Tendon-driven designs, like the Red Sea Exploratorium hand and Ocean One hands, show promise by utilizing kinematic coupling and preloaded elastic joints to enhance underwater performance [23, 24]. However, achieving versatile grasping, maintaining robustness in dynamic water flows, withstanding high water pressure, and achieving waterproof remain challenging for finger-based hand designs.

Interestingly, underwater creatures have not developed multi-digit manipulation to the same degree as their terrestrial counterparts. This highlights the potential of non-finger-based grippers for effective underwater grasping and manipulation, which has been successfully validated by some robotics trial [25, 26]. Many fish, lacking arms and legs, rely solely on their mouths for complex feeding and eating tasks, demonstrating agile and effective interaction with underwater environments [27].

Inspired by the fish mouth feeding process, we propose a non-finger-based gripper solution to the challenges of robotic grasping and manipulation in underwater environments. Our alternative approach introduces the fish mouth inspired origami gripper, leveraging the Yoshimura origami pattern, requiring only one Degree of Freedom to perform robust grasping tasks underwater. The origami gripper offers scalable design, grasping compliance, and robustness, enabling pinch, power grasp, grasping multiple objects simultaneously, and scooping from the seabed. Through the fusion of origami and robotics, we present the design, modeling, fabrication, and validation of the origami gripper, demonstrating its potential for enhancing robotic grasping and manipulation in underwater environments. The paper proceeds with sections detailing the concept of the origami gripper, concluding with insights on its promising underwater applications.

The remaining structure of the article is as follows: In Section II, the design and fabrication of the origami gripper inspired by fish mouths is presented. Section III covers the experimental validation and characterization of the origami gripper's fundamental grasping performance. Section IV showcases the actual underwater grasping performance of the origami gripper, followed by the conclusion and future work.

## II. DESIGN AND FABRICATION

### A. Design of Origami Gripper

The design of the origami robot hand is based on the Yoshimura origami pattern [28-30], as shown in Fig. 2(A_left). The process involves transforming a 2D Yoshimura pattern into a 3D thin-walled structure (Fig. 2(A_middle)) by folding and securing the lateral sides with driven ears, thereby creating a basic origami gripper (Fig. 2(A_right)). This design exploits the mechanical phenomenon of buckling in thin-walled cylinders under compression. While previous studies have demonstrated the Yoshimura pattern's effectiveness in creating chamber-like actuators that generate motion through changes in length and shape [31, 32], limited research has explored its deformation and folding capabilities for robotic gripper design. In this study, constraints were introduced to the original Yoshimura structure to convert its expanding motions into gripping-like motions induced by a uniaxial pulling force.

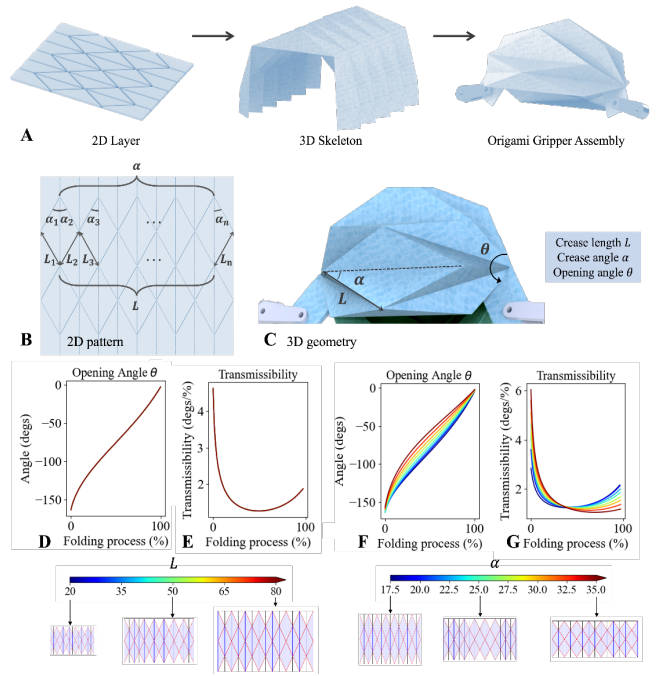

**Figure 2.** Design of the Origami gripper. (A) 2D to 3D folding and assembly. (B) Major design parameters for Yoshimura Origami pattern. (C) Assembled origami gripper with three major parameters. Opening angle (D) and transmission ratio (E) as a function of folding process for varying L. Opening angle (F) and transmission ratio (G) as a function of folding process for varying α.

The deformation of the origami hand is governed by two key parameters: crease angles $\alpha$ and cease lengths $L$. The vector $\alpha = [\alpha_1, \alpha_2, \alpha_3, \cdots, \alpha_n]$ represents the angles between intersecting crease lines, and $L = [L_1, L_2, L_3, \cdots, L_n]$ is a vector of lengths between intersecting points, as illustrated in Fig. 2(B). The opening angle $\theta \in [-\pi, 0]$ determines the folding and unfolding of the origami hand, where $\theta = -\pi$ indicates that the origami hand is fully open (flat fold) and $\theta = 0$ indicates a fully closed position.

To investigate the impact of these parameters on the origami robot hand. We applied geometric approach to simulate the forward kinematics of origami robot hands with diverse Yoshimura crease patterns [33]. By iteratively determining vertex positions through geometry intersections, we gained insights into the folding motion intricately tied to specific patterns. The parameter L was varied uniformly between 20 mm and 80 mm while α was held constant at 25°, resulting in simulations of the opening angle θ and transmission ratio during the hand's closing motion.

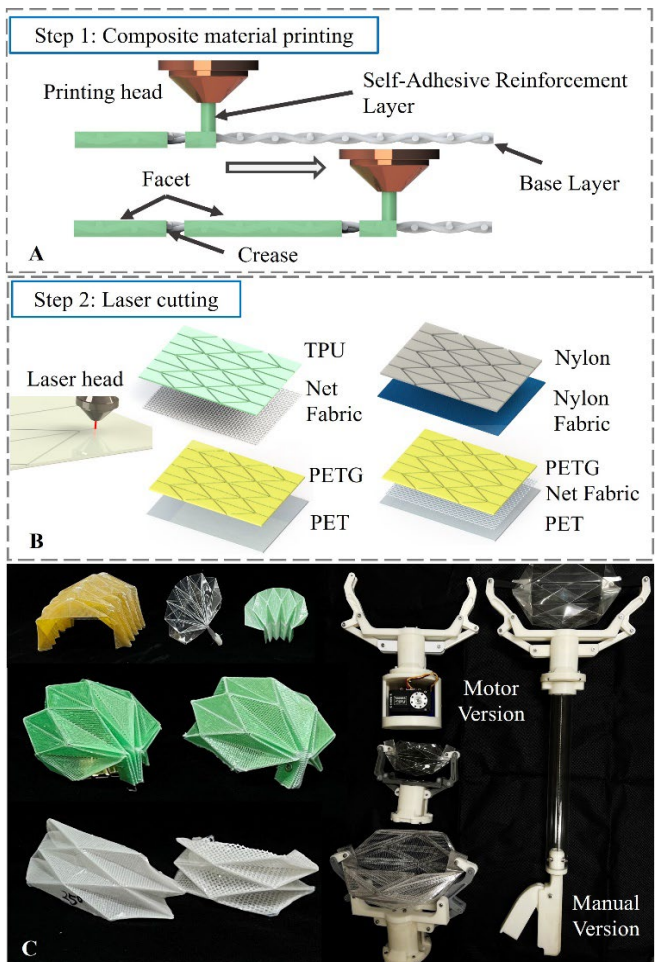

**Figure 3** Fabrication process of the origami gripper. (A) Involves the composite material 3D printing of a 2D origami sheet. (B) Shows the laser cutting of the origami sheet with the desired folding patterns. (C) Displays the prototype origami grippers after folding. These grippers can be mounted on a motor base or a manual base for grasping purposes.

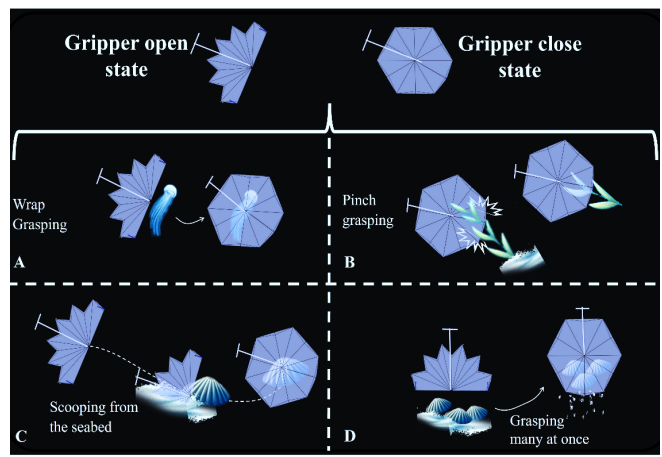

**Figure 4.** Multi-type grasping of origami gripper. (A) Wrap grasping. (B) Pinch grasping. (C) Scooping from the seabed. (D) Grasping many objects at once.

Subsequently, α was varied between 17.5° and 35° while keeping L constant, and the simulations were repeated (Fig. 2C, D). Due to variations in the scales of origami crease patterns, x-axis measurements were normalized to represent the percentage of total folding, ranging from 0% to 100% between θ = [-180°, 0°]. The results indicate that varying L does not significantly affect the folding process characteristics, as all origami crease patterns exhibit similar trajectories and transmissibility, demonstrating scalability in design. Conversely, altering α results in distinct trajectories and transmissibility, offering potential advantages for diverse grasping applications.

### B. Fabrication of Origami Gripper

The fabrication process of the origami gripper involves two primary steps, starting with the Single Layer Printing of Composite Sheet Preparation. Traditional materials like paper or single-layer sheets lack the durability needed for repeated folding and unfolding due to weak crease lines and facets. To address this issue, a composite sheet is utilized, offering flexibility at the creases while enhancing strength and maintaining rigidity at the facets. The composite sheet consists of a base layer and a top self-adhesive reinforcement layer, as depicted in Fig. 3(A). The base layer, typically constructed from materials like fabric or PET, provides the necessary flexibility and strength at the creases. On the other hand, the top reinforcement layer, made of thermoplastic materials, adheres to the base layer when heated and solidifies upon cooling to enhance facet rigidity. The fabrication of the composite sheet involves fixing the base layer and adhering the top layer to the facet regions using additive manufacturing techniques. This selective reinforcement ensures that only the facets gain additional rigidity. Both the base and top layers can be composed of various materials, with some examples listed for reference.

After the composite sheet preparation, laser cutting is employed to achieve the desired folding patterns, guided by the crease pattern from the top reinforcement layer, leading to the creation of a 3D origami structure. Different combinations of composite sheets are used to fabricate various origami grippers, as depicted in Fig. 3(C). These grippers are then paired with an actuation base, available in motorized and manual versions, to enhance durability and functionality for diverse applications.

### C. Multiple Grasping Types of Origami Gripper

While the proposed origami gripper operates with a single actuation degree of freedom for transitioning between open and closed states as shown in Fig. 4, it offers versatility in achieving four distinct grasping types by adjusting its position and interacting with the environment [34, 35].

**Type 1: Wrap Grasping** - The gripper can fully envelop a target smaller than its chamber as shown in Fig.4 (A), providing a robust and gentle grip, ensuring stability when handling delicate objects. This mode is particularly effective in creating a secure enclosure for grasping slippery,

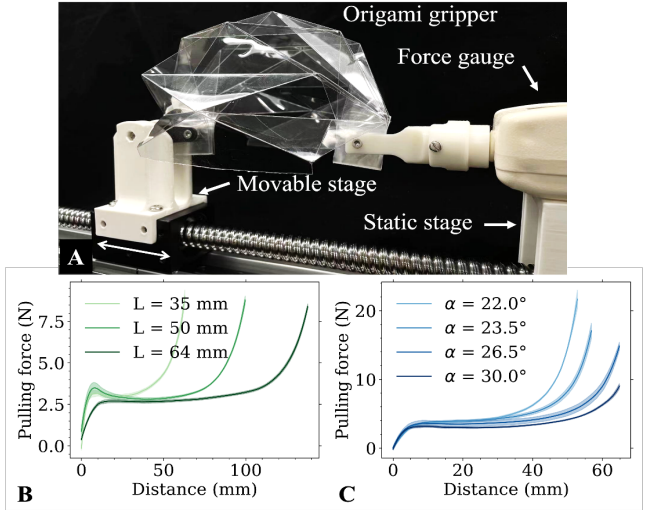

**Figure 5** Uniaxial pulling experiments. (A) Experimental setup. (B) The relationship between pulling force and pulling distance for robot hands with different crease angles $\alpha$. (C) Experimental pulling force as a function of pulling distance curves for robot hands with different crease lengths $L$.

irregularly shaped, soft, deformable, or dynamically moving targets.

**Type 2: Pinch Grasping** - By using its end edge to pinch objects, the gripper can achieve precise pinch grasping as shown in Fig. 4(B), ideal for grasping small or slender objects that require a specific point of contact.

**Type 3: Scooping** - Equipped with a self-contained chamber, the gripper can scoop objects such from the seabed as shown in Fig. 4(C), facilitating the collection of partially or fully buried items and creatures.

**Type 4: Grasping Multiple Objects Simultaneously** - The gripper can perform regional grasping, capturing multiple targets within its grasp space in a single motion as shown in Fig. 4(D). This capability enhances efficiency in collecting underwater objects and creatures compared to traditional finger-based methods, allowing for the retrieval of multiple items at once.

### III. EXPERIMENTS

In the experiments, we characterized the fundamental performance of the origami gripper for grasping. Three key tests were conducted: actuation force and displacement measurement, grasping robustness through a pull-out test with a grasped ball, and a pinch force test.

#### A. Uniaxial Actuation Force and Actuation Displacement Experiment

To comprehend the force transmission from input to output, we conducted an in-depth examination of the relationship between pulling force and displacement. By fixing one handle and quasi-statically pulling the other away on a testbed, we observed a positive correlation between pulling force and handle displacement (Fig. 5A). During the closing process, it exhibited an initial decrease followed by an increase in stiffness with a handle extension. A plateau region exists at phase II, indicating a stage when the origami robot hand is at its neutral state, and a nearly constant pulling force drives the origami robot hand opening at the plateau region. After this stage, stretching the two ends further requires significant changes in the pulling force to overcome the material elastics. Given the origami pattern's flexibility concerning crease length and crease angles (Fig. 2B), we conducted systematic experiments with origami robot hands featuring different crease patterns by altering two key parameters: crease angle $\alpha$ and cease length L. As L increased from 35 mm to 64 mm (Fig. 5B), the force-displacement relationships displayed similar behaviors, with variations in actuation distance. As $\alpha$ increases from 22° to 30° a critical reduction in actuation stiffness was observed (Fig. 5C). Origami robot hands with larger $\alpha$ are observed to be actuated using less pulling force at the closing stage.

#### B. Grasped Ball Pulling Out Experiment

To assess the grasping robustness of the origami robot hand, we measured the maximum pull-out force required to quasi-statically remove an object grasped by the robot hand using envelope grasping (Fig. 6A). For simplicity, we employed spherical objects, with diameters of 40mm, 50mm, and 63mm, as testing objects. Each spherical object was firmly held by the origami robot hand in a consistent manner. A string was then securely attached to each spherical object, with the other end of the string connected to a force gauge. By applying a controlled pulling force to the string, we determined the maximum force required for each spherical object to slip out of the origami robot hand's grasp. Experimental results concerning the pull-out force (Fig. 6B) revealed a trend where the pull-out force increased with the diameter of the objects. In the case of smaller objects, the robot hand's internal structure often left substantial space during grasping, with the fingertip sections providing the primary supporting force. Consequently, these sections exhibited relative flexibility, allowing objects to escape more easily. Conversely, larger objects tended to fit within the robot hand snugly, enabling the entire origami body to generate significant supporting force. This indicates that the robot hand can hold heavy objects by choosing a proper hand size. Beyond origami crease parameters, the pull-out force and holding capability should also depend on the materials, which is the future work of this research.

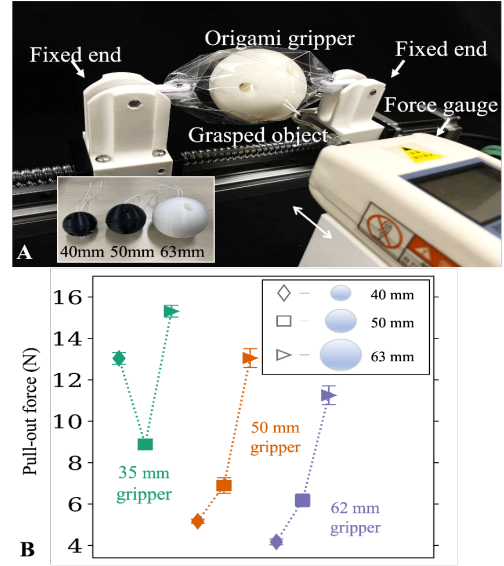

**Figure 6.** Grasping robustness of grasped ball pulling out experiment. (A) Experimental setup. (B) Testing result.

## C. Origami Gripper Pinch Force Experiment

The deployment process effectively converted the uniaxial pulling force into a closing motion and gripping force. Pinch force was defined as the normal force exerted between the fingertip and target objects in the direction of quasi-static fingertip closing as characterized by the platform shown in Fig. 7(A). Fig. 7(B, D, F) illustrate an increase in pinch force as the handles are pulled. The pinch force is influenced by origami crease parameters, which determine the geometry of the robot hand and impact the transmission of motion and force from input to output. The sensitivity of pinch force was investigated (refer to Fig. 7C, E, G), and defined as the ratio of the changes in output pinch force to a change in pulling distance. Increasing α within a certain range initially increases and then decreases pinch force sensitivity. Higher sensitivity indicates that less pulling distance is needed to generate the same level of output pinch force. The highest sensitivity was observed at the crease pattern with a crease angle near 28.5°, suggesting that origami robot hands can be tailored for different applications.

## D. Demonstration of the Origami Robot Hand

The underwater grasping performance of the origami gripper was evaluated in both controlled laboratory setups and real underwater environments. As illustrated in Fig. 8(A1-A2), a manual handle version of the origami gripper was deployed at an underwater diving site near Pattaya Island for the collection of sea urchins. The unique origami design allows the entire grasping system to operate without the need for waterproofing considerations, even when handling spinose objects such as sea urchins. A specialized version of the origami gripper, featuring a hybrid layer with air-permeable facets, was used to conduct the onsite grasping.

In Fig. 8B, a wrap grasping technique is demonstrated on fish being transferred from the sea to the deck of a boat and then into a fish barrel. The distinctive design of the origami gripper enables it to securely encase the fish, even when they struggle and jump, minimizing the risk of slippage and dropping.

In Fig. 8C, a pinch grasp is applied to a shallow water sea urchin, showcasing both robustness and adaptability. Fig. 8D illustrates a scooping grasp at the shallow water seabed, where the gripper collects a partially buried rhopilema esculentum. The gripper effectively scoops the rhopilema while allowing small sand particles to pass through the air holes, making it highly efficient for collecting objects buried in the sand.

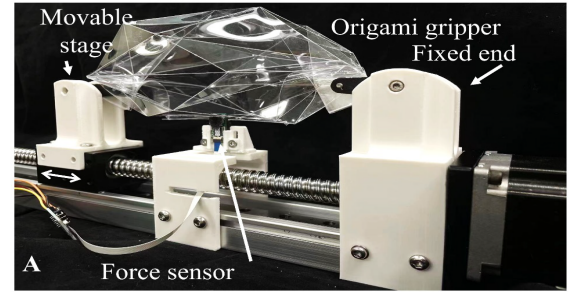

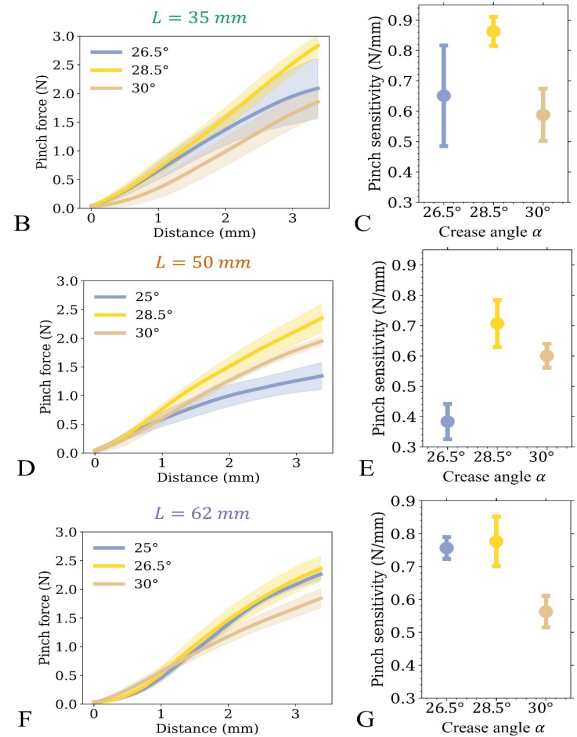

**Figure 7.** Pinch grasping performance characterization. (A) Experimental setup. (B, D, F) The relationship between pinch force and pulling distance. (C, E, G) Experimental pinch force sensitivity as a function of crease angle.

Fig. 8(E1-E3) demonstrates the gripper's ability to handle a jellyfish, an extremely delicate and vulnerable creature. The gripper safely wraps around the jellyfish without applying contact pressure, ensuring its integrity.

Fig. 8(F1-F3) shows the effective grasping of a crab, a spinous and irregularly shaped creature, by the origami gripper, which firmly grasps it.

Fig. 8(G1-G3) demonstrates the collection of two abalones simultaneously, highlighting the gripper's high efficiency and accuracy in grasping tasks.

Furthermore, the prototype origami gripper has successfully completed over 300 underwater grasping tasks, demonstrating robust performance and reliable integration with the grasping target and environment. This showcases the preliminary durability and effectiveness of the origami gripper design.

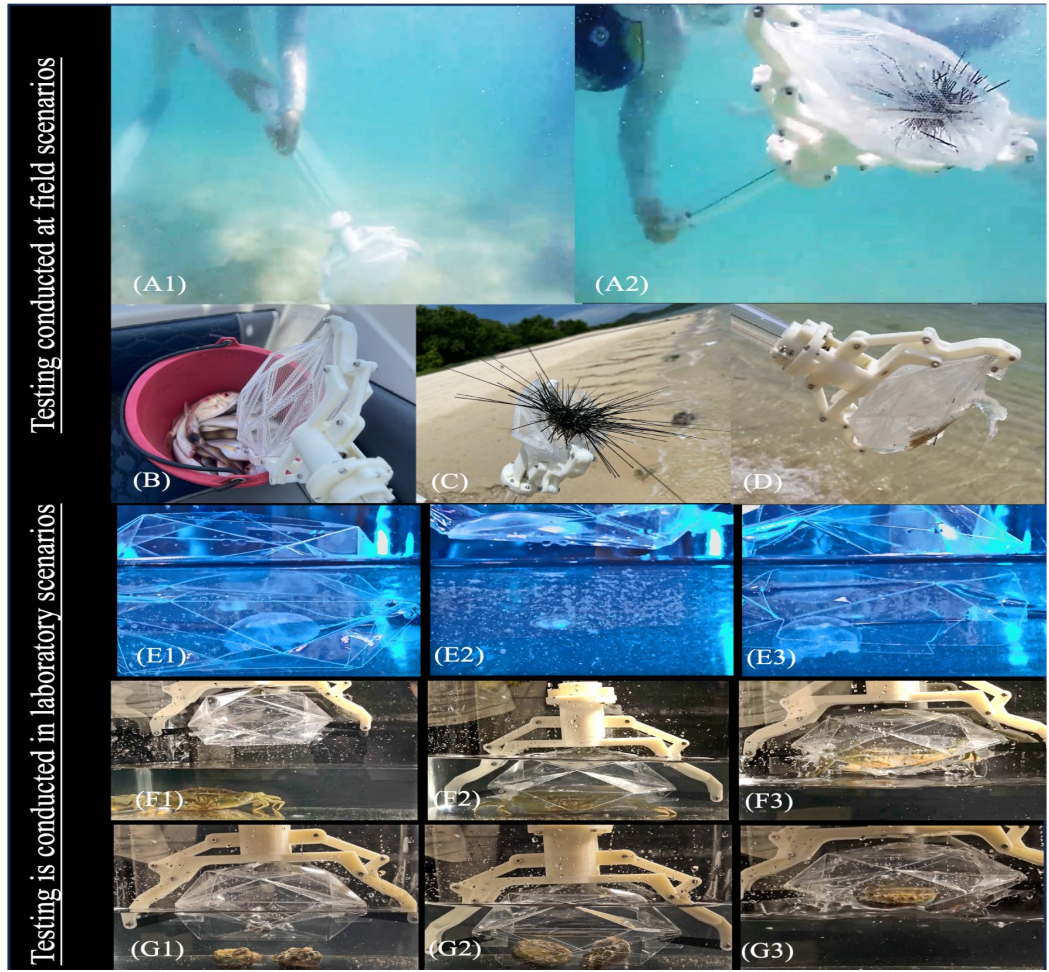

**Figure. 8.** Grasping demonstrations. (A1-A2) Underwater sea urchin collection. (B) Fish collection from deck. (C) Shallow water sand dollar collection. (D) Shallow water rhopilema esculentum collection. (E1-E3) Water tank jelly fish collection. (F1-F3) Water tank crab collection. (G1-G3) Water tank abalones collection.

## IV. VIII. Conclusion and Future Work

This paper introduces a fish mouth-inspired origami gripper as an innovative solution for robust underwater grasping tasks, requiring only one dof to achieve multi-type grasping. Leveraging the Yoshimura origami pattern, this gripper offers scalability, compliance, and robustness for various grasping actions in underwater environments. The paper emphasizes the importance of marine exploration, addressing the challenges in underwater sampling and manipulation. It explores the effectiveness of underactuated grippers, soft robotic hands, and hydraulic continuum-style fingers in underwater applications, setting the stage for the origami gripper's unique approach to underwater robotics. Design, fabrication, modeling, and dedicated characterization validate the gripper's efficacy, showcasing excellent grasping robustness and efficiency in a laboratory setting and real underwater environments across four typical grasping modes targeting various underwater creatures. This work represents a promising advancement in the development of gripping tools tailored for underwater grasping, demonstrating the origami gripper's potential in enhancing underwater exploration and research.

Future work could resolve further refining the design and functionality of the origami gripper to enhance its performance in underwater applications. Potential avenues include optimizing the origami pattern for improved grasping capabilities, exploring advanced materials to enhance durability and pressure resistance, and integrating sensor technologies for precise control and feedback mechanisms. Additionally, investigating autonomous operation and adaptive grasping strategies could expand the gripper's utility in complex underwater scenarios. Collaborative efforts with marine biologists and environmental scientists may also offer valuable insights into specific underwater tasks suitable for the origami gripper, broadening its applications in marine exploration and research.